\pdfoutput=1

\documentclass[11pt]{article}

\usepackage[final]{ACL2023}

\usepackage{times}
\usepackage{latexsym}
\usepackage{amsmath}
\usepackage{amssymb}
\usepackage{booktabs}
\usepackage{tabularx}
\usepackage{graphicx}
\usepackage{xcolor,colortbl}
\usepackage{xspace}
\usepackage{mathtools, nccmath}
\usepackage{soul}
\usepackage{enumitem}
\usepackage[utf8]{inputenc}
\usepackage{csquotes}
\usepackage[capitalize]{cleveref}
\usepackage{anyfontsize}

\usepackage[T1]{fontenc}

\usepackage[activate={true,nocompatibility}, kerning=true,spacing=true, stretch=30,shrink=30]{microtype}
\microtypecontext{spacing=nonfrench}

\usepackage{inconsolata}

\definecolor{yellow}{RGB}{251,188,4}
\definecolor{mgreen}{rgb}{0.20, 0.66, 0.33}
\definecolor{green}{RGB}{32,33,36}
\definecolor{red}{RGB}{165, 14, 14}
\definecolor{blue}{RGB}{23,78,166}
\definecolor{orange}{RGB}{227,116,0}
\definecolor{forestgreen}{rgb}{0.13, 0.55, 0.13}
\definecolor{background}{HTML}{ccff99}

\newcommand{\taskname}{\text{MEI}}

\newcommand{\be}{\mathbf{e}}
\newcommand{\bm}{\mathbf{m}}
\newcommand{\mcE}{\mathcal{E}}
\newcommand{\mcP}{\mathcal{P}}
\newcommand{\mcM}{\mathcal{M}}
\newcommand{\mcD}{\mathcal{D}}
\newcommand{\hatM}{\mcM_\text{other}}

\newcommand{\allM}{\mcM_\text{all}}
\newcommand{\modelname}{MEIRa}
\newcommand{\longformer}{\phi}
\newcommand{\meanstd}[2]{#1{\scriptsize{$\pm$}#2}}

\renewcommand{\paragraph}[1]{\vspace{1mm}\noindent\textbf{#1}}

\definecolor{errorblue}{RGB}{0,114,178}
\definecolor{ForestGreen}{RGB}{34,139,34}

\newcommand{\hlcyan}[1]{{\sethlcolor{cyan!50}\hl{#1}}}
\newcommand{\hlyellow}[1]{{\sethlcolor{yellow!50}\hl{#1}}}
\newcommand{\hlgreen}[1]{{\sethlcolor{ForestGreen!50}\hl{#1}}}
\newcommand{\hlgrey}[1]{{\sethlcolor{black!20}\hl{#1}}}
\newcommand{\hlred}[1]{{\sethlcolor{red!20}\hl{#1}}}
\newcommand\Mark[1]{\textsuperscript#1}

\makeatletter
\DeclareRobustCommand\onedot{\futurelet\@let@token\@onedot}
\def\@onedot{\ifx\@let@token.\else.\null\fi\xspace}

\def\eg{\textit{e.g}\onedot} 

\def\ie{\textit{i.e}\onedot}

\def\vs{\textit{vs}\onedot}

\makeatother

\title{Major Entity Identification: \\
A Generalizable Alternative to Coreference Resolution}

\author{Kawshik Manikantan{\normalfont \Mark{1}}, Shubham Toshniwal{\normalfont \Mark{2}}, Makarand Tapaswi{\normalfont \Mark{1}}, Vineet Gandhi{\normalfont \Mark{1}}\\
\Mark{1}CVIT, IIIT Hyderabad \quad \Mark{2}NVIDIA\\
\small{ kawshik.manikantan@research.iiit.ac.in, stoshniwal@nvidia.com, \{makarand.tapaswi, vgandhi\}@iiit.ac.in }
  }
 
\begin{document}
\maketitle

\begin{abstract}

The limited generalization of coreference resolution (CR) models has been a major bottleneck in the task's broad application. 
Prior work has identified annotation differences, especially for mention detection, as one of the main reasons for the generalization gap and proposed using additional annotated target domain data. 
Rather than relying on this additional annotation, we propose an alternative referential task,
\textbf{M}ajor \textbf{E}ntity \textbf{I}dentification (\taskname),
where we:
(a)~assume the target entities to be specified in the input, and
(b)~limit the task to only the frequent entities. 
Through extensive experiments, we demonstrate that MEI models generalize well across domains on multiple datasets with supervised models and LLM-based few-shot prompting. 
Additionally, MEI fits the classification framework, which enables the use of robust and intuitive classification-based metrics. 
Finally, MEI is also of practical use as it allows a user to search for all mentions of a particular entity or a group of entities of interest.
\footnote{Code for the paper is available at \url{https://github.com/KawshikManikantan/MEI}}

\end{abstract}

\section{Introduction}
\label{sec:intro}

Coreference resolution (CR) is the task of finding text spans that refer to the same entity. 
CR is a fundamental language understanding task relevant to various downstream NLP applications, such as 
question-answering~\cite{dhingra-etal-2018-neural},
building knowledge graphs~\cite{koncel-kedziorski-etal-2019-text}, and 
summarization~\cite{sharma-etal-2019-entity}.
Despite the importance of CR and the progress made by neural coreference models~\cite{dobrovolskii-2021-word, bohnet-etal-2023-coreference,zhang-etal-2023-seq2seq}, domain generalization remains an issue even with the best-performing supervised models~\cite{xia-van-durme-2021-moving, toshniwal-etal-2021-generalization}.

\begin{figure}[t!]
\centering
\includegraphics[width=0.99\linewidth]{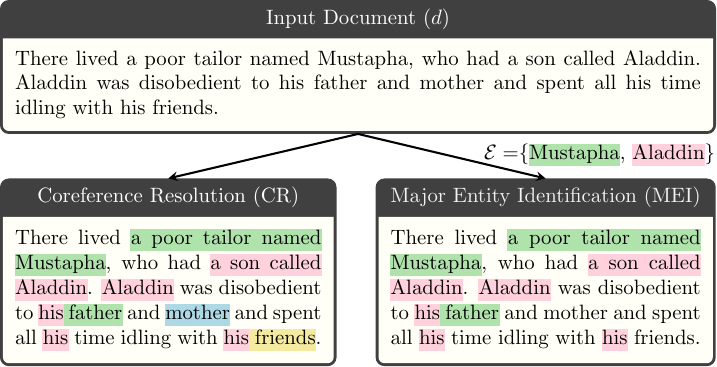}
\caption{CR \vs~MEI. 
The CR task aims to detect and cluster all mentions into different entities, shown in various colors.
MEI takes major entities as additional input and aims to detect and classify the mentions that refer only to these entities.
} 
\label{fig:teaser}
\end{figure}

The lack of domain generalization in CR models can largely be attributed to differences in annotation guidelines of popular CR benchmarks, specifically annotation guidelines about what constitutes a mention~\cite{porada2023investigating}.
For example, OntoNotes~\cite{pradhan-etal-2013-towards} does not annotate singletons, confounding mention identity with being referential.
Thus, models trained on OntoNotes generalize poorly~\cite{toshniwal-etal-2021-generalization}.  
The importance of mention detection for CR generalization is further highlighted by~\citet{gandhi-etal-2023-annotating}, showing that solely annotating mentions is sufficient and more efficient for adapting pre-trained CR models to new domains (in comparison to annotating coreference chains). 
Similarly, GPT-4 struggles with zero-/few-shot mention prediction, but with ground-truth mentions, its CR performance~\cite{le2023large} is competitive with that of supervised models~\cite{toshniwal-etal-2021-generalization}.

Given these observations, we hypothesize that current CR models, including large language models, generalize well at \textit{mention clustering} but struggle to generalize on \textit{mention detection} due to idiosyncrasies of different domains/benchmarks. We put forth an alternative referential task where the entities of interest are known and provided as additional input. Assuming entities to be part of the input offloads the required domain adaptation from training to inference. Specifically, we propose the task of Major Entity Identification (\taskname{}), where we assume the major entities of the narrative, to be provided as input along with the text (see \cref{fig:teaser}). We focus on major entities for the following reasons:
(a)~Specifying major entities of a narrative is intuitively easier. 
(b)~A handful of major entities often dominate any discourse. \cref{tab:dataset} shows that in FantasyCoref roughly 16\% of entities (942 of 5829) contribute to 63\% of the mentions (35938 of 56968).




In this work, we adapt two literary CR benchmarks, namely LitBank~\cite{bamman-etal-2020-annotated} and FantasyCoref~\cite{han-etal-2021-fantasycoref} by identifying frequently occurring entities as major entities and customizing a state-of-the-art coreference model~\cite{toshniwal-etal-2021-generalization} to MEI. 
Our tests for generalizability reveal that while there is a big gap in CR performance between in- and out-of-domain models~\cite{toshniwal-etal-2021-generalization}, this performance gap is much smaller for MEI (Section~\ref{sec:exp_sup_models}). 
To test this hypothesis further, we evaluate large language models (LLMs) for MEI in a few-shot learning setup. 
On CR, LLMs are shown to struggle with mention detection and perform worse than supervised models~\cite{le2023large}.
Contrary to this, on MEI, top LLMs (\eg~GPT-4) are only slightly behind supervised models (Section~\ref{sec:exp_llms}).
These experiments in the supervised setting and the few-shot setting demonstrate that the MEI task is more generalizable 
than CR. 

\begin{table}[t]
\small
\tabcolsep=0.15cm %
\centering
\begin{tabular}{l cc cc}
\toprule
& \multicolumn{2}{c}{LitBank} & \multicolumn{2}{c}{FantasyCoref} \\ 
Statistics & CR & MEI & CR & MEI \\
\midrule
\# of Mentions & 29103 & 16985 & 56968 & 35938 \\
\# of Non singletons & 23340 & 16985 & 56968 & 35938 \\
Mean ant. dist. & 55.31 & 36.95 & 57.58 & 30.24 \\
\midrule
\# of Clusters & 7927 & 490 & 5829 & 942 \\
Avg. cluster size & 3.67 & 34.66 & 9.77 & 38.15 \\
\bottomrule
\end{tabular}
\caption{Comparing CR and MEI. MEI has fewer but larger clusters, and a smaller mean antecedent distance (Mean ant. dist.). Our formulation's frequency-based criterion for deciding major entities means that singleton mentions are typically not a part of MEI.
}
\label{tab:dataset}
\end{table}

Additionally, we argue that MEI is easier to evaluate than CR.
The MEI task can be viewed as a classification task in which any text span either refers to one of the input entities or the null class (\textit{minor} entities and other non-mention spans).
The classification metrics maintain consistent granularity, proportionally penalize perturbations, and exhibit high discriminatory power while intuitively meeting multiple desired specifications~\citep{moosavi-strube-2016-coreference,10.1017/S135132491000029X}.


Furthermore, MEI, by its definition, disregards insignificant and smaller clusters known to inflate the CR metrics~\cite{moosavi-strube-2016-coreference, lu-ng-2020-conundrums, kummerfeld-klein-2013-error}. As an aside, formulating MEI as a classification task allows for a trivial parallelization across candidate spans (Appendix~\ref{sec:app_speed_comp}).

Finally, MEI's explicit mapping of mentions to predefined entities improves its usability over CR in downstream applications that focus on mentions of specific entities. MEI effectively replaces tailored heuristics employed to extract CR cluster(s) referring to entities of choice in such applications
(entity understanding~\cite{inoue-etal-2022-learning},
sentiment and social dynamics analysis~\cite{zahiri2017emotion, Antoniak_2023}).

%
%

\section{Task Formulation}
\label{sec:task_defn}
\paragraph{Notation.}
For a document $d$, let $\mcE = \{e_j\}_{j=1}^L$ be the set of $L$ major entities that we wish to identify.
We define $\allM$ as the set of all mentions that could refer to any entity and subsequently $\mcM_j \subseteq \allM$ as the set of mentions that refer to a major entity $e_j$.
Furthermore, we denote $\mcM = \bigcup_j \mcM_j$ as the set of mentions that refer to one of the major entities while mentions that do not correspond to any major entity are designated as  $\hatM = \allM \setminus \mcM$.

\paragraph{Task formulation.}
In MEI, the input consists of the document $d$ and designative phrases $\mcP = \{p(e_j)\}_{j=1}^L$ where
$p(e_j)$ succinctly represents the entity $e_j$.  
For example, in \cref{fig:teaser}, the phrases \emph{``Aladdin''} and \emph{``Mustapha''} uniquely represent Aladdin and his father who appear in \emph{``Aladdin And The Wonderful Lamp''}.
Note that in CR, the designative phrases $\mcP$ are not part of the input.

In contrast to CR's clustering foundations, MEI starts with a prior for each entity (the designative phrase) and can be formulated as an open set classification, where every mention is either classified as one of the major entities or ignored.
Formally, MEI aims to assign each mention $m \in \mcM_j$ to $e_j$ and mentions $m \in \hatM$ to $\varnothing$, a null entity.

\section{Supervised MEI models}
\label{sec:supervised}

We propose \modelname, \textbf{M}ajor \textbf{E}ntity \textbf{I}dentification via \textbf{Ra}nking, which draws inspiration from the entity ranking formulation~\cite{xia-etal-2021-incremental,toshniwal-etal-2020-learning} and maintains an explicit representation for entities.
The \modelname~models consist of 3 steps: encoding the document, proposing candidate mentions, and an identification (id) module that tags mentions with major entities or the null entity.

\paragraph{Document encoding}
is performed using a Longformer-Large~\cite{beltagy2020longformer}, $\longformer$, that we finetune for the task.
Mentions (or spans) are encoded as $\bm_i = \longformer(m_i, d)$ by concatenating the first, last, and an attention-weighted average of the token representations within the mention span.
In MEI, an additional input is the set of designative phrases $\mcP$ for the major entities. Since each phrase is derived from the document itself, we also obtain its encoding using the backbone: $\be_j = \longformer(p(e_j),d)$.

\paragraph{Mention detection.}
Similar to prior efforts~\cite{toshniwal-etal-2021-generalization}, we use a mention proposal network that predicts high-scoring candidate mentions.
This step finds all mentions $\allM$ and not just the ones corresponding to the major entities $\mcM$.Training a model to only detect mentions of major entities would confuse it leading to poor performance.

\begin{figure*}[t]
\centering
\includegraphics[width=0.9\textwidth]{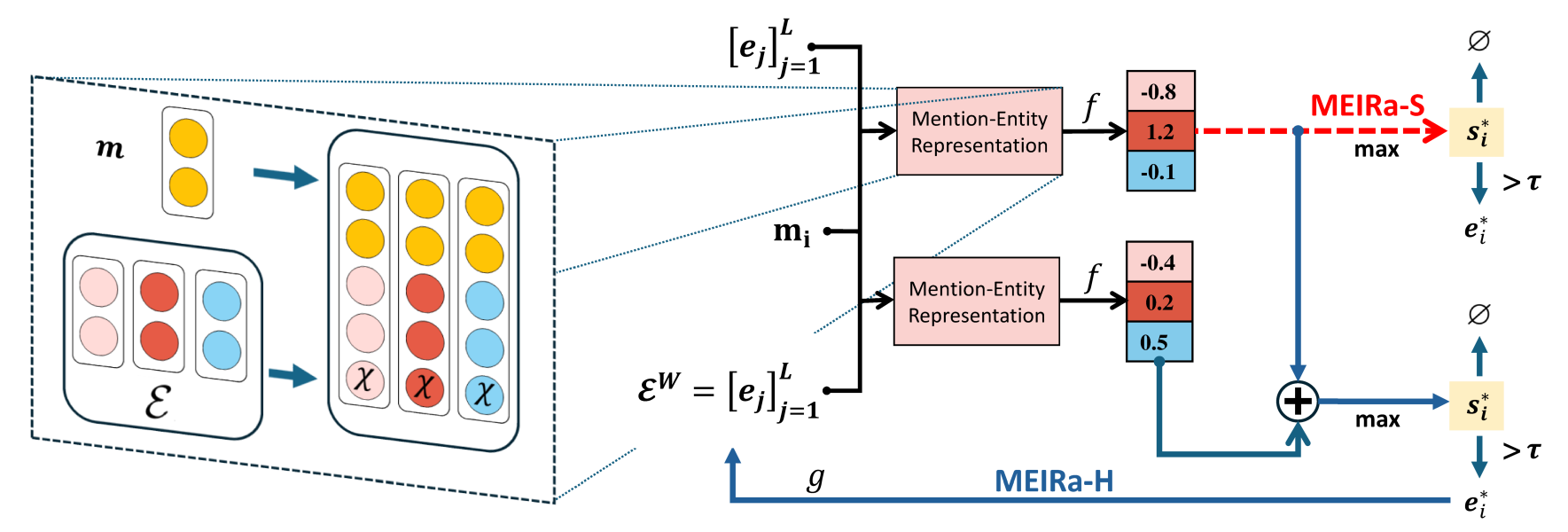}
\caption{Identification module of \modelname.
A mention encoding $\bm_i$ is concatenated with each entity's embedding in $\mcE^W$ and the metadata $\chi(m_i, e_j)$.
Network $f$ scores the likelihood of assigning $m_i$ to each major entity.
If the highest score $s_i^*$ is above the threshold $\tau$, $m_i$ is associated with the highest scoring major entity $e_i^*$ or discarded.
In \modelname-S, the entity memory $\mcE^W$ remains static. For \modelname-H (blue path), the assigned entity's working memory is updated, and both the static (top half) and updated working memory (bottom half) are utilized to compute a final score.}
\label{fig:method}
\end{figure*}

\paragraph{Identification module.}
As illustrated in \cref{fig:method}, 
we initialize a working memory $\mcE^W = [\be_j]_{j=1}^L$ as a list of $L$ major entities based on their designative phrase representations.
Given a mention $m_i$, the id module computes the most likely entity as:
\begin{equation}
 [s_i^*,  e_i^*] = \max_{j{=}1\ldots L} \ f([\bm_i, \be_j, \chi(m_i, e_j)]) \, ,
\end{equation}
where $f()$ is an MLP that predicts the score of tagging mention $m_i$ with the entity $e_j$, and $\chi(m_i, e_j)$ encodes metadata.
The output $s_i^*$ corresponds to the highest score and $e_i^*$ is the top-scoring entity.
Based on the score, $m_i$ is assigned to:
\begin{equation}
y(m_i) = 
\begin{cases}
e_i^* & \text{ if } s_i^* > \tau \, , \\
\varnothing & \text{ otherwise } \, ,
\end{cases}
\end{equation}
where $\tau$ is a threshold (set to 0 in practice).

The metadata $\chi(m_i, e_j)$ contains a distance (position) embedding representing the log distance between the mention $m_i$ and the last tagged instance of the entity $e_j$.
If no mention is yet associated with the entity, we use a special learnable embedding.

\paragraph{Updates to the working memory.}
We investigate two approaches:

(i)~\textbf{\modelname-S}tatic:
As the name suggests, the working memory $\mcE^W$ of the entity representations remains constant ($\mcE^{W(0)}$) and is not updated with new mention associations. This makes the approach highly parallelizable.

(ii)~\textbf{\modelname-H}ybrid: Similar to traditional CR, this variation maintains a dynamic working memory $\mcE^W$, which is updated with every new mention-id association. Specifically, assuming $m_i$ is assigned to $e_j^*$, the working memory would be updated using a weighted mean operator $g$ as $\be_j \leftarrow g(\be_j, \bm_i)$, similar to~\citet{toshniwal-etal-2020-learning}. To prevent error accumulation, we evaluate the mentions against $\mcE^W$ and the initial entity representations ($\mcE^{W(0)}$), then compute the average score.
This hybrid approach reaps benefits from both, the initial clean designative phrases 
and the dynamic updates. 

Following \citet{toshniwal-etal-2020-learning}, the mention detection and identification modules are trained end-to-end using separate cross-entropy loss functions.

\section{Few-shot MEI with LLMs}

\label{sec:few-shot-llms}

We propose a prompting strategy to leverage LLMs for MEI, addressing their challenges in CR.

\paragraph{Mention detection challenges.}
CR or MEI can be addressed using separate few-shot prompting strategies for mention detection and mention clustering/identification. However,~\citet{le2023large} found that this strategy faced significant challenges with mention detection, performing worse than a deterministic mention detector.
Thus, they assume access to an oracle mention detector and focus on evaluating LLMs' linking capabilities.

An alternative is to use an external supervised mention detector instead of the oracle. However, this requires annotated training data and may not align with a true few-shot LLM prompt paradigm. 
Additionally, supervised mention detectors often fail to generalize across CR datasets due to annotation variability~\cite{lu-ng-2020-conundrums}.

\paragraph{MEI with LLMs.}
We demonstrate that transitioning from CR to MEI addresses this gap in mention detection and proposes an end-to-end, few-shot prompting approach for MEI. 
Inspired by~\citet{dobrovolskii-2021-word}, we develop a prompting strategy that first performs MEI at word-level (rather than span), followed by a prompt to retrieve the span corresponding to the word.

In addition to the document $d$ and the set of phrases $\mcP$, we also provide entity identifiers (\eg~\#1, \#2) to the LLM.
We will use the following example:

\noindent%
\underline{Document:}
\textit{That lady in the BMW is Alice's mom.}\\
\underline{Major Entities}:
1.~\textit{Alice};
2.~\textit{Alice's mother}.

\paragraph{Prompt 1. Word-level MEI.}
Mention detection with LLMs is challenging due to the frequent occurrence of nested mentions.
We overcome this by prompting the LLM to tag each word.
Specifically, through few-shot examples, 
we ask the LLM to detect and tag the \textbf{syntactic heads}%
\footnote{A syntactic head of a phrase is a word (\textit{lady}) that is central to the characteristics of the phrase ({\textit{The lady in the BMW}}).}
(e.g., \textit{lady}, \textit{Alice}, \textit{mom}) of mentions that refer to the major entities.
Other words are left untagged (implicitly assigned to $\varnothing$, the null entity).
To create the few-shot examples, a contiguous set of words annotated with the same entity is considered as a span and its syntactic head is extracted using \texttt{spaCy}~\cite{Honnibal_spaCy_2020}.

The ideal output for the example above is:

\textit{``That lady\#2 in the BMW is Alice\#1's mom\#2..''}.

Note that, even though the span \textit{``BMW''} might be a valid mention, it is not annotated as it does not refer to one of the major entities.
The exact prompt used for this is provided in the Appendix, \cref{tab:app_wl}.

\paragraph{Prompt 2. Head2Span retrieval.}
The entity tagged heads are passed to the Head2Span (H2S) module, along with the document to retrieve the span.
The prompt consists of the document pre-annotated with the positions of the head, where each candidate head-word is followed by a ``\#'' and is instructed to be replaced by the complete span (including any existent determiners and adjectives).
For the input:

\textit{That lady\# in the BMW is Alice\#'s mom\#.}

\noindent the expected ideal output is

\textit{That lady (That lady in the BMW) in the BMW is Alice(Alice's)'s mom (Alice's mom).}

Table~\ref{tab:app_h2s} in the appendix shows the H2S prompt.

\paragraph{Preserving structure.}
We pose MEI as a structured generation task, prompting LLMs to reproduce documents and generate MEI tags at specific locations.
Proprietary models like GPT-4 generally reproduce documents faithfully but for rare failures, we use the Needleman-Wunsch algorithm~\cite{needleman1970general} to align documents and extract tags
In the case of open-source models, we employ regular expression-based constrained decoding with the \texttt{outlines} library~\cite{willard2023efficient} 

\section{Experiments}
\label{sec:experiments}
\paragraph{Datasets.} We evaluate three literary datasets chosen for their longer length and identifiable major entities, particularly the key narrative elements such as characters or plot devices. \cref{tab:dataset} compares statistical aspects of MEI and CR, revealing that MEI features fewer clusters (entities) but larger cluster sizes (more mentions per cluster).

(i)~\emph{LitBank}~\cite{bamman-etal-2020-annotated} annotates coreference in 100 literary texts, each averaging around 2000 words. Following prior work~\cite{toshniwal-etal-2021-generalization}, we utilize the initial cross-validation split, dividing the documents into training, validation, and test sets with an 80:10:10 ratio.

(ii)~\emph{FantasyCoref}~\cite{han-etal-2021-fantasycoref} provides OntoNotes~\cite{pradhan-etal-2013-towards}-style\footnote{The exact guidelines are documented  \href{https://www.ldc.upenn.edu/sites/www.ldc.upenn.edu/files/english-coreference-guidelines.pdf}{here}} coreference annotations for 211 documents from Grimm’s Fairy Tales, with an average length of approximately 1700 words. The dataset includes 171 training, 20 validation, and 20 test documents.

(iii)~\emph{Additional Fantasy Text ($AFT$)} ~\cite{han-etal-2021-fantasycoref} provides annotations for long narratives:
(a)~Aladdin (6976 words),
(b)~Ali Baba and the Forty Thieves  (6911 words), and
(c) Alice in Wonderland (13471 words).

\paragraph{Metrics.}
In contrast to CR, MEI facilitates the use of simple classification metrics. We define standard precision and recall for each major entity considered as an individual class of its own.

For a dataset $\mcD = \{d_1, \ldots, d_{|\mcD|}\}$, the evaluation metrics are defined as follows:
\begin{equation}
\small
\text{Macro-F1} = \frac{\sum_{d \in \mcD} \sum_{e_j \in \mcE_d} F1(e_j)} {\sum_{d \in \mcD} | \mcE_d|} \,, \, \text{and}
\end{equation}
\begin{equation}
\small
\text{Micro-F1} = \frac{1}{|\mcD|}
\sum_{d \in \mcD} \frac{\sum_{e_j \in \mcE_d} F1(e_j) \cdot |\mcM_j|} { \sum_{e_j \in \mcE_d} |\mcM_j|} \, .
\end{equation}
Macro-F1 is the average F1-score of entities across the dataset, while Micro-F1 is the frequency-weighted F1-score of entities within a document, averaged across the dataset.

\paragraph{Major entity selection.} 
We select as major entities, the top-$k$ entities ranked as per the frequency of occurrences. 
We use $k{=}5$ for LitBank and FantasyCoref after visualizing the frequency plots of their training sets. 
For longer documents in AFT, we select up to 9 entities to ensure coverage of all key entities from the story.
We also enforce that every entity $e_j \in \mcE$ has a mention count $|\mcM_j| \geq 5$. We derive the representative span for each selected $e_j$ from the set of mentions $\mcM_j$ by selecting the most commonly occurring name or nominal mention.

\paragraph{Implementation details.} \mbox{ }

\noindent \textit{Supervised models}: 
Model hyperparameters are derived from~\citet{toshniwal-etal-2021-generalization}.
To ensure consistent performance across different numbers of target entities, we randomly select a subset of major entities at each training iteration (more details in \cref{app:var_ent}). Supervised models were trained five times with random seeds, and we present aggregated results as the mean and standard deviation.

\begin{table}[t]
\centering
\footnotesize
\tabcolsep=0.1cm
\begin{tabular}{@{}lcccc@{}}
\toprule
\multicolumn{1}{l}{} & \multicolumn{2}{c}{FantasyCoref} & \multicolumn{2}{c}{LitBank} \\ 
Model & Macro-F1 & Micro-F1 & Macro-F1 & Micro-F1 \\
\midrule
Coref-ID & \meanstd{72.5}{2.2} & \meanstd{78.8}{2.7} & \meanstd{79.7}{2.7} & \meanstd{80.6}{3.7} \\
Coref-CM & \meanstd{77.7}{1.8} & \meanstd{82.4}{2.2} & \meanstd{74.1}{2.5} & \meanstd{76.0}{3.0} \\
Coref-FM & \meanstd{77.9}{1.7} & \meanstd{83.2}{2.2} & \meanstd{77.4}{2.3} & \meanstd{80.6}{4.7} \\
\midrule
\modelname-S & \textbf{\meanstd{80.7}{0.6}} & \textbf{\meanstd{84.9}{0.5}} & \meanstd{80.8}{0.8} & \meanstd{81.8}{1.0} \\
\modelname-H & \meanstd{80.3}{1.4} & \meanstd{84.3}{2.0} & \textbf{\meanstd{82.3}{1.2}} & \textbf{\meanstd{83.2}{2.5}} \\
\bottomrule
\end{tabular}
\caption{Results for models trained jointly on FantasyCoref and LitBank.}
\label{tab:supervised}
 \vspace{-0.1in}
\end{table}

\noindent \textit{LLMs:}
We follow a few-shot prompting mechanism across the setups and experiments. 
Prompts that perform referential tasks consist of 3 examples of 6 sentences each.
These 3 examples contain a mixture of narrative styles (narratives, dialogues), types of entities (major, non-major entities), categories of mentions (names, nominals, pronouns), and plurality.
Additionally, before producing the MEI output, we ask the LLM to describe each major entity briefly.
We find that this additional step improves performance.
For the H2S prompt, we provide 9 sentences as examples, balancing the number of pre- and post-modifiers to the head.
All examples were selected from LitBank's train set and kept constant throughout the experiments.
We set the temperature to 0 for all the models to ensure consistent and reproducible outputs.

\subsection{Experiments: Supervised Models}
\label{sec:exp_sup_models}
\paragraph{Baselines.} 
We train the \texttt{longdoc} model~\cite{toshniwal-etal-2021-generalization} for CR and perform the following three inference-time adaptations for MEI:

\emph{Coref-ID:}
\texttt{longdoc} uses active lists of entity representations, resolving coreference by associating mentions with existing clusters or generating new ones.
During inference, we disable the cluster creation step and pre-fill the entity list with the encoded vector representations of the major entities. Hence, all the detected mentions either get mapped to one of the major entities or are discarded.

\emph{Coref-Cosine Map} (Coref-CM): Since CR clusters obtained from \texttt{longdoc} lack explicit entity association, we employ the Kuhn-Munkres (KM) algorithm~\cite{munkres1957algorithms} to find the optimal matching cluster for each major entity.
The cost matrix uses the cosine similarity between the encoded representation of the major entities and the predicted cluster embeddings, both derived from \texttt{longdoc}.

\emph{Coref-Fuzzy Map} (Coref-FM):
This method uses the KM algorithm to derive optimal mappings by constructing a cost matrix from accumulated fuzzy-string matching scores between designative phrases and the predicted cluster's mention strings.

\begin{table}[t]
\centering
\footnotesize
\tabcolsep=0.1cm
\begin{tabular}{@{}lcccc@{}}
\toprule
\multicolumn{1}{l}{} & \multicolumn{2}{c}{FantasyCoref} & \multicolumn{2}{c}{LitBank} \\ 
Model & Macro-F1 & Micro-F1 & Macro-F1 & Micro-F1 \\ \midrule
Coref-ID & \meanstd{63.4}{1.8} & \meanstd{69.5}{3.6} & \meanstd{58.0}{2.4} & \meanstd{57.7}{1.0} \\
Coref-CM & \meanstd{72.8}{0.3} & \meanstd{76.5}{0.5} & \meanstd{61.0}{5.9} & \meanstd{61.2}{5.2} \\
Coref-FM & \meanstd{71.2}{1.5} & \meanstd{75.2}{1.3} & \meanstd{66.1}{2.1} & \meanstd{67.1}{3.9} \\
\midrule
\modelname-S & \textbf{\meanstd{75.7}{1.5}} & \meanstd{78.5}{1.2} & \meanstd{74.6}{1.1} & \meanstd{74.7}{1.6} \\ 
\modelname-H & \meanstd{74.7}{1.0} & \textbf{\meanstd{78.5}{0.8}} & \textbf{\meanstd{77.2}{1.9}} & \textbf{\meanstd{78.6}{2.7}} \\
\bottomrule
\end{tabular}
\caption{Results for models trained on OntoNotes.}
\label{tab:generalisation}
\vspace{-0.1in}
\end{table}

\paragraph{Supervised results.}
In this experiment, we train \modelname~and the baseline models on the joint training set of LitBank and FantasyCoref. 
Subsequently, we assess their performance on the individual test sets, with results summarized in \cref{tab:supervised}.
Overall, \modelname~models consistently outperform the baselines on both metrics while also exhibiting better stability with a lower variance. The considerable variance observed in the performance of baseline methods across all experiments underscores the non-trivial nature of identifying clusters corresponding to major entities within the output clusters provided by the CR algorithms. \modelname-H and \modelname-S exhibit competitive parity on FantasyCoref (children stories), while \modelname-H edges out on LitBank dataset, showcasing its adaptability in elaborate sentence constructions.

\paragraph{Generalization across datasets.}
To evaluate the generalization capabilities of \modelname~and baseline models, we train them on the OntoNotes dataset and then test their performance on LitBank and FantasyCoref.
The results are presented in \cref{tab:generalisation}.
When compared with \cref{tab:supervised}, we observe a significant performance drop across the baseline models (\eg~for Coref-ID, the average Micro-F1 scores drop from 80.6 to 57.7 on LitBank).
The performance gap for the baseline models is more pronounced on LitBank than on FantasyCoref because LitBank's annotation strategies differ more significantly from those of OntoNotes.  
The observations aligns with previous work~\cite{toshniwal-etal-2021-generalization}, that showcase poor generalization of models trained for CR.
In contrast, \modelname~models recover most of the underlying performance on both the datasets (\modelname-H drops a little from 83.2 to 78.6 on LitBank Micro-F1), demonstrating MEI as a more adaptable task, bringing robustness over varying annotation strategies.

\begin{table}[t]
\centering
\footnotesize
\begin{tabular}{@{}lcc@{}}
\toprule
\multicolumn{1}{l}{} & \multicolumn{2}{c}{AFT} \\ 
Model & Macro-F1 & Micro-F1 \\ \midrule
Coref-ID & \meanstd{68.1}{5.9} & \meanstd{78.7}{6.1} \\
Coref-CM & \meanstd{71.1}{2.8} & \meanstd{82.4}{4.2} \\
Coref-FM & \meanstd{71.1}{4.7} & \meanstd{83.2}{4.7} \\
\midrule
\modelname-S & \meanstd{81.6}{1.4} & \meanstd{88.8}{1.3} \\
\modelname-H & \textbf{\meanstd{82.8}{1.1}} & \textbf{\meanstd{89.5}{1.0}} \\ \bottomrule
\end{tabular}
\caption{Results on the AFT dataset.}
\label{tab:long}
\end{table}

\paragraph{Long documents.} 
\cref{tab:long} presents results on the AFT dataset of the models trained using a combined training set of LitBank and FantasyCoref.
\modelname~models significantly outperform the baseline models, with \modelname-H~gaining 11.7\% in Macro-F1 over the best baseline. The results demonstrate the efficacy of \modelname~models on resolving key entities in longer narratives.

\paragraph{Computational performance.} 
\modelname-S supports parallel batched processing since it does not update the working memory after associating mentions, \ie~the mentions need not be processed sequentially from left to right. 
Hence, post-mention detection (common to all models), \modelname-S is about 25$\times$ faster than \texttt{longdoc} when assessed across LitBank, FantasyCoref and AFT datasets on an NVIDIA RTX 4090 (see~\cref{fig:speed_comp} in the appendix). 
Additionally, with the model's small memory footprint during inference, the entire process can also be parallelized across chunks of documents making it extremely efficient.
Hence, we pose \modelname-S as a faster while competitive alternative to \modelname-H (that requires dynamic updates and has similar computational performance as \texttt{longdoc}).

\begin{table}[t]
\footnotesize
\tabcolsep=0.04cm
\begin{tabular}{@{}lcccc@{}}
\toprule
\multicolumn{1}{l}{} & \multicolumn{2}{c}{FantasyCoref} & \multicolumn{2}{c}{LitBank} \\ 
Model & Macro-F1 & Micro-F1 & Macro-F1 & Micro-F1 \\ \midrule
\modelname-H & 88.5 & 91.0 & 86.1 & 85.4 \\\midrule
GPT-4 & {\textbf{90.7}} & \textbf{92.0} & {\textbf{88.8}} & {\textbf{91.6}} \\
GPT-3.5 & 69.2 & 74.2 & 74.3 & 75.8 \\\midrule
Code Llama-34B &  67.0 & 72.4 & 68.9 & 73.1 \\
Llama3-8B & 53.8 & 60.6 & 50.2 & 53.4 \\
Mistral-7B & 67.3 & 75.8 & 61.6 & 73.9 \\
\bottomrule
\end{tabular}
\caption{ Few-shot LLM prompting results assuming the availability of ground-truth mentions.
}
\label{tab:llm_setup2}
\end{table}

\subsection{Experiments: Few-shot prompting}
\label{sec:exp_llms}
\paragraph{Models.}
We experiment with GPT-4\footnote{Specifically, \texttt{gpt-4-1106-preview}}~\cite{openai2024gpt4}, GPT-3.5\footnote{Specifically, \texttt{gpt-3.5-turbo-0125}}, Code Llama-34B ~\cite{roziere2023code}, Mistral-7B ~\cite{jiang2023mistral}, and Llama3-8B.\footnote{\href{https://ai.meta.com/blog/meta-llama-3/}{https://ai.meta.com/blog/meta-llama-3/}}
Following \citet{le2023large}, we use the instruction-tuned versions for open-source models.
These models were chosen for their ability to handle the extended context required for our benchmarks.

\subsubsection{Linking Performance w/ Gold Mentions} 
We first evaluate all the models assuming the availability of an oracle mention detector.
The experimental configuration is aligned with that of~\citet{le2023large}, albeit with the distinction that we assess them for the MEI task rather than for CR.
The prompt used in our setup is provided in \cref{tab:app_link} of Appendix. 
For comparison, we also perform inference on golden mentions with \modelname-H.

The results in \cref{tab:llm_setup2} show that GPT-4 surpasses the supervised \modelname-H model in this setup. 
Among LLMs, GPT-4 is easily the best-performing model. 
Code Llama-34B performs the best among open-source models, closely followed by Mistral-7B.  
While Code Llama-34B is tailored for the code domain, surprisingly, it outperforms strong LLMs suited for natural language. 
This result corroborates a similar finding by ~\citet{le2023large} for CR and related evidence regarding code pretraining aiding entity tracking ~\cite{kim2024code}.
We find that Code Llama-34B performs close to GPT-3.5 for FantasyCoref, though a sizable gap persists in the Macro-F1 metric for LitBank , potentially due to its linguistic complexity.

\subsubsection{MEI Task Performance with LLMs}

In this section, we present the results for the end-to-end MEI task using LLMs. We compare all the approaches from Section~\ref{sec:few-shot-llms} and relevant baselines with the results summarized in  \cref{tab:llm_setup4}. 
To limit the combinations of LLMs and approaches for our experiments, we first compare all the approaches in tandem with GPT-4 and then present results for the best-performing approach with other LLMs. 

The first straightforward approach of using a \emph{Single Prompt} to retrieve all the mentions of major entities in a single pass results in a significant performance drop compared to \modelname-H (prompt in \cref{tab:app_dir} of Appendix). The reason is that while GPT-4 outperforms \modelname-H at mention linking, its mention detection performance, especially with nested mentions, is much worse compared to \modelname-H.\footnote{The failure to detect nested mentions is despite best efforts to provide illustrative examples in the few-shot prompt. \citet{le2023large} report similar findings with earlier GPT versions.
}

\begin{table}[t]
\centering
\footnotesize
\tabcolsep=0.04cm
\begin{tabular}{@{}lcccc@{}}
\toprule
\multicolumn{1}{l}{} & \multicolumn{2}{c}{FantasyCoref} & \multicolumn{2}{c}{LitBank} \\ 
Model & Macro-F1 & Micro-F1 & Macro-F1 & Micro-F1 \\ \midrule
\modelname-H & \textbf{80.3} & \textbf{84.3} & 82.3 & 83.2 \\ 
GPT-4 w/ Ext det & 80.1 & 82.2 & 78.7 & 83.9 \\
\midrule
\multicolumn{5}{@{}l}{\textbf{GPT-4 with varying prompting strategies}} \\
Single prompt & 51.8 & 57.5 & {61.1} & {70.7} \\
Two-stage prompt &  70.5 &  74.9 &  76.5 &  81.3 \\
\midrule
\multicolumn{5}{@{}l}{\textbf{Word-level MEI + \texttt{spaCy} H2S}} \\
GPT-4 & 77.1 & 79.4 & \textbf{82.5} & \textbf{85.5} \\
GPT-3.5 & 50.1 & 54.4 & 60.1 & 63.1 \\
Code Llama-34B & 30.0 & 31.4 & 22.7 & 23.2 \\
Llama3-8B & 29.2 & 32.1 & 20.5 & 26.0 \\
Mistral-7B & 19.4 & 21.9 & 12.9 & 14.0 \\
\bottomrule
\end{tabular}
\caption{Results on LLMs with different mention detection and linking strategies.}
\label{tab:llm_setup4}
\end{table}

To further underscore the importance of mention detection, we compare against \emph{GPT-4 w/ Ext det}, which utilizes an external pre-trained mention detector followed by prompt-based linking (prompt in \cref{tab:app_link} of Appendix).
We train the mention detector on the PreCo dataset~\cite{chen-etal-2018-preco}, which achieves a 93.8\% recall and 53.1\% precision on the combined FantasyCoref and LitBank validation sets. \emph{GPT-4 w/ Ext det} performs at par with the fully supervised \modelname-H, again highlighting the strong mention linking capabilities of GPT-4. 

Next, we present the results of our proposed \emph{Two-stage prompt}, motivated by the \emph{Single prompt} method's failure with nested mentions. The first prompt asks GPT-4 to perform word-level MEI, by limiting the task to syntactic heads only. The second prompt then performs the task of mapping the identified syntactic heads to full mention spans. The results strongly validate our proposed approach with a relative improvement of more than 10\% over the \emph{Single prompt} method across all metrics and datasets. 
We also explore replacing the second step, i.e., head-to-span (H2S) retrieval, with an external tool. Specifically, we invert \texttt{spaCy}'s span-to-head mapping to obtain a head-to-span retriever.\footnote{For the test set gold mentions of the two datasets, there were only two cases where spans had the same head. We handled these two cases manually.}

\begin{table}[t]
\centering
\footnotesize
\tabcolsep=0.1cm
\begin{tabular}{@{}lcc@{}}
\toprule
\textbf{Error Type} & \textbf{\modelname-H} & \textbf{GPT-4} \\ \midrule
Missing Major & 162 & 793 \\
Major-Major & 210 & 154 \\
Major-Other & 243 & \phantom{11}0 \\
Other-Major & 200 & 516 \\
Extra-Major & 461 & 896 \\ \midrule
Total & 1276 & 2359 \\\bottomrule
\end{tabular}%
\caption{Breakdown of errors by \modelname-H and GPT-4 on the combined LitBank and FantasyCoref test set.}
\label{tab:err_analysis}
\end{table}

GPT-4 significantly improves in this setup, outperforming even the supervised model on LitBank.
Given the strong performance of \emph{GPT-4 + spaCy H2S}, we evaluate the open-source LLMs in only this setting. 
We observe a wide gap between GPT-4 and the open-source models.
Llama3-8B outperforms other open-source models on both datasets in Micro-F1 and stays competitive with the larger Code Llama-34B in Macro-F1. However, this contrasts with Llama3-8B's significant lag in the idealized golden mention setting, which solely evaluates the model's linking capabilities.


\subsection{Error Analysis}

\begin{table}[]

\centering
\scriptsize{
\setlength{\tabcolsep}{2pt}
    \centering
    \begin{tabular}{p{0.08\textwidth}  p{0.37\textwidth}}
    \toprule
  \textbf{Golden Mentions}   &  
  
  Presently \textcolor{errorblue}{[a small boy]$_0$} came walking along the path -- \textcolor{errorblue}{[an urchin of
nine or ten]$_0$} \ldots \ldots \textcolor{errorblue}{[Winterbourne]$_1$} had immediately perceived that \textcolor{errorblue}{[he]$_1$}
might have the honor of claiming \textcolor{errorblue}{[him]$_2$} as a fellow countryman. ``Take
care \textcolor{errorblue}{[you]$_2$} don't hurt \textcolor{errorblue}{[your]$_2$} teeth," \textcolor{errorblue}{[he]$_1$} said, paternally \ldots \ldots \textcolor{errorblue}{[My]$_2$}
mother counted them last night, and one came out right afterwards. She
said she'd slap \textcolor{errorblue}{[me]$_2$} if any more came out. \textcolor{errorblue}{[I]$_2$} can't help it. It’s this old
Europe \ldots \ldots If \textcolor{errorblue}{[you]$_2$} eat three lumps of sugar, \textcolor{errorblue}{[your]$_2$} mother will
certainly slap \textcolor{errorblue}{[you]$_2$}," \textcolor{errorblue}{[he]$_1$} said. ``She's got to give \textcolor{errorblue}{[me]$_2$} some candy,
then," rejoined \textcolor{errorblue}{[[his]$_1$ young interlocutor]$_2$}.
 \\\midrule
 
\textbf{GPT-4  \phantom{00} Output}       
&  
Presently \textcolor{errorblue}{[a small boy]$_0$} came walking along the path -- \textcolor{errorblue}{[an urchin of
nine or ten]$_0$} \ldots \ldots \textcolor{errorblue}{[Winterbourne]$_1$} had immediately perceived that \textcolor{errorblue}{[he]$_1$}
might have the honor of claiming \textcolor{errorblue}{[him]$_2$} as a fellow countryman. ``Take
care \textcolor{red}{\hlyellow{you}} don't hurt \textcolor{red}{\hlyellow{your}} teeth," \textcolor{errorblue}{[he]$_1$} said, paternally \ldots \ldots \textcolor{errorblue}{[My]$_2$}
mother counted them last night, and one came out right afterwards. \textcolor{red}{\hlcyan{[She]$_2$}}
said \textcolor{red}{\hlcyan{[she]$_2$}}'d slap \textcolor{errorblue}{[me]$_2$} if any more came out. \textcolor{errorblue}{[I]$_2$} can't help it. \textcolor{red}{\hlgrey{[It]$_2$}}’s this old
Europe \ldots \ldots If \textcolor{red}{\hlyellow{you}} eat three lumps of sugar, \textcolor{errorblue}{[your]$_2$} mother will
certainly slap \textcolor{errorblue}{[you]$_2$}," \textcolor{errorblue}{[he]$_1$} said. ``\textcolor{red}{\hlcyan{[She]$_2$}}'s got to give \textcolor{errorblue}{[me]$_2$} some candy,
then," rejoined \textcolor{red}{\hlgreen{[his]$_2$}\hlyellow{ young interlocutor}}.
\\\midrule

\textbf{MEIRa-H Output}       &  

Presently \textcolor{red}{\hlyellow{a small boy}} came walking along the path -- \textcolor{red}{\hlred{[an urchin of
nine or ten]}} \ldots \ldots \textcolor{errorblue}{[Winterbourne]$_1$} had immediately perceived that \textcolor{errorblue}{[he]$_1$}
might have the honor of claiming \textcolor{errorblue}{[him]$_2$} as a fellow countryman. ``Take
care \textcolor{errorblue}{[you]$_2$} don't hurt \textcolor{errorblue}{[your]$_2$} teeth," \textcolor{errorblue}{[he]$_1$} said, paternally \ldots \ldots \textcolor{errorblue}{[My]$_2$}
mother counted them last night, and one came out right afterwards. She
said she'd slap \textcolor{errorblue}{[me]$_2$} if any more came out. \textcolor{errorblue}{[I]$_2$} can't help it. It’s this old
Europe \ldots \ldots If \textcolor{errorblue}{[you]$_2$} eat three lumps of sugar, \textcolor{errorblue}{[your]$_2$} mother will
certainly slap \textcolor{errorblue}{[you]$_2$}," \textcolor{errorblue}{[he]$_1$} said. ``She's got to give \textcolor{errorblue}{[me]$_2$} some candy,
then," rejoined \textcolor{errorblue}{[[his]$_1$ young interlocutor]$_2$}.
\\ \bottomrule

    \end{tabular}
}
    \caption{Qualitative Analysis showcasing different errors made by GPT-4 and \modelname-H. Errors are color-coded as follows: \textcolor{black}{\hlyellow{Missing Major}}, \textcolor{black}{\hlcyan{Others-Major}}, \textcolor{black}{\hlgrey{Extra-Major}}, \textcolor{black}{\hlgreen{Major-Major}}, and {\hlred{Major-Other}}.
}
    \label{tab:QA}
\end{table}

We classify MEI errors into five categories: 
(1)~\emph{Missing Major:} Not detecting a mention $m \in \mcM$.
(2)~\emph{Major-Major:} Assigning a mention $m \in \mcM_j$ to any other major entity $\mcE \setminus e_j$.
(3)~\emph{Major-Other:} Assigning a mention $m \in \mcM$ to $\varnothing$. 
(4)~\emph{Other-Major:} Assigning a mention $m \in \hatM$ to any major entity in $\mcE$.
(5)~\emph{Extra-Major:} Detecting extra mentions $m \not \in \allM$ and assigning to any major entity in $\mcE$.

Results combined over the LitBank and FantasyCoref test sets are presented in \cref{tab:err_analysis}. Missing Major and Extra-Major contribute most of the errors for GPT-4, highlighting the scope for improvement in mention detection and span retrieval. Mention detection also remains a challenge in \modelname-H, the model making most of the mistakes in the Extra-Major category. 
GPT-4 distinguishes major entities more clearly than \modelname-H but tends to over-associate other mentions with major entities, resulting in higher Other-Major and Extra-Major errors.
Note that GPT-4 has zero errors in the Major-Other category due to the prompt design, which only allows annotating major entities. Examples of these errors are visualized in \cref{tab:QA}.

\section{Related Work}
\paragraph{Neural models for CR} have become the \textit{de facto} choice in supervised settings~\cite{lee-etal-2017-end,kantor-globerson-2019-coreference, joshi-etal-2020-spanbert,otmazgin-etal-2023-lingmess}. 
Efforts to enhance model efficiency include reducing candidate mentions to word-level spans~\cite{dobrovolskii-2021-word} and using single dense representations for entity clusters~\cite{xia-etal-2021-incremental,toshniwal-etal-2020-learning}.

\paragraph{Generalization in CR} remains a lingering problem~\cite{moosavi-strube-2017-lexical,zhu-etal-2021-ontogum, porada2023investigating}. Current solutions include  
feature addition~\cite{aralikatte-etal-2019-rewarding,otmazgin-etal-2023-lingmess},
joint training~\cite{xia-van-durme-2021-moving,toshniwal-etal-2021-generalization}, and active learning~\cite{zhao-ng-2014-domain,yuan-etal-2022-adapting,gandhi-etal-2023-annotating}. Rather than relying on additional training data, we argue for an alternative formulation where the burden of domain adaptation is offloaded from training to inference.

\paragraph{LLM evaluation on referential tasks}
has largely been conducted in limited settings, such as the sentence-level  Winograd Schema Challenges (WSC)~\cite{brown2020language}, clinical pronoun resolution~\cite{agrawal-etal-2022-large} and instance-level Q\&A~\cite{yang-etal-2022-gpt}. \citet{le2023large} conducted the first document-level evaluation of LLMs for CR but assumed an oracle-mention detector. In contrast, we conduct end-to-end evaluations.


\paragraph{Entity-centric tasks} similar to MEI include character identification, where either annotations are restricted to a subset of entities~\cite{baruah-narayanan-2023-character} or custom models are developed to extract mentions of specific characters from TV show transcripts~\cite{chen-choi-2016-character,zahiri2017emotion}.
We differ from these works by adopting a generalized task formulation independent of annotation strategies and entity selection. Another task, {Entity Linking} \cite{Ji2015OverviewOT} extracts distinct entities from a document and links them to external Knowledge Bases. In contast, MEI focuses on retrieving all mentions (including nominals and pronominals) of a specific set of key entities, extracted solely from the context of the document. 

\section{Conclusion}
CR models are limited in their generalization capabilities owing to annotation differences and general challenges of domain adaptation. 
We propose \taskname{} as an alternative to CR, where the entities relevant to the input text are provided as input along with the text.  Our experiments demonstrate that MEI is more suited for generalization than CR. Additionally, MEI can be viewed as a classification task that enables the use of intuitive metrics. A trivially parallelized variation (\modelname-S), gives a 25x speedup over a comparable CR model, making it more suitable for longer narratives. Unlike CR, the formulation of \taskname{} allows few-shot prompted LLMs to effectively compete with trained models. Our novel two-stage prompting and robust baseline methods empower top-performing LLMs like GPT-4 to achieve this. Our analysis indicates that this task holds promise for effectively evaluating the long-context referential capabilities of LLMs in an end-to-end manner.

Potential applications of MEI include domains such as film and literature, where metadata about salient entities can be sourced from external databases like IMDb or SparkNotes. Additionally, MEI can be applied to the analysis of documents like of financial and legal reports, when the user is familiar with the relevant entities. Lastly, recent research~\cite{lin-zeldes-2024-gumsley} indicates that LLMs can assist or automate the extraction of salient entities, a direction we intend to explore in future work. 




\section{Limitations}
Major Entity Identification (MEI) is proposed as a generalizable alternative to the coreference resolution (CR) task, and is not a replacement of CR. MEI limits itself to major entities and only caters to applications that are interested in a particular pre-defined set of entities.  Our experiments follow certain thresholds that might not be universally applicable, and results and performance might vary slightly along this decision (refer \cref{app:var_ent}). Our current few-shot prompting evaluations are limited only to a few models that accommodate a large context window. Optimizing prompts and architecture to allow for a piece-wise aggregation of outputs across chunks of documents is left for future work.

\bibliography{custom}

\bibliographystyle{acl_natbib}

\appendix
\clearpage
\section{Appendix}
\label{sec:appendix}

\subsection{Linking Speed Comparison}
\label{sec:app_speed_comp}
\begin{figure}[h!]
\centering
\includegraphics[width=0.99\linewidth]{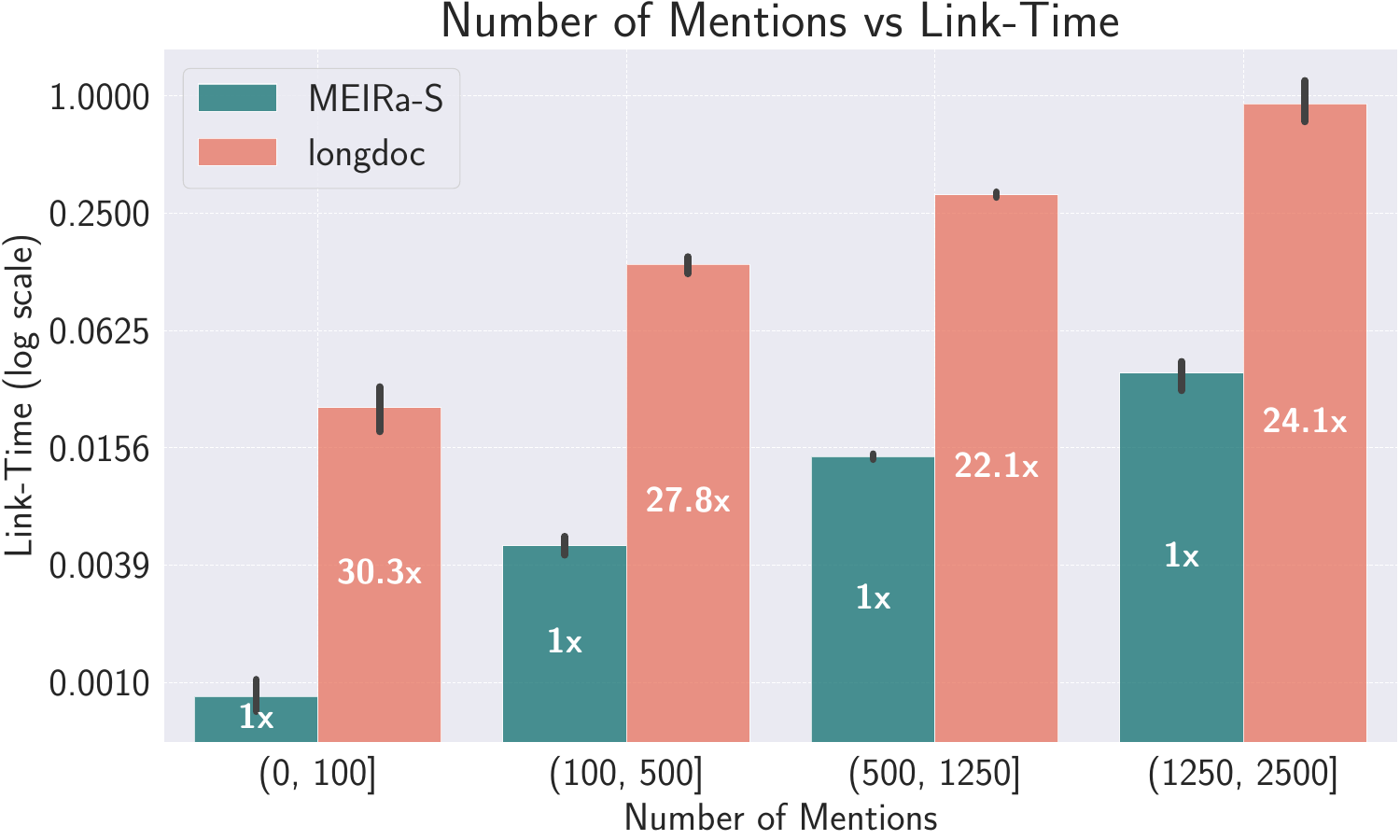}
\caption{Linking speed comparison between ~\modelname-S and \texttt{longdoc} for the combined LitBank and FantasyCoref test set.
There exists 6 documents with (0, 100] mentions, 19 with (100, 500] mentions, 5 with (500, 1250] mentions and 3 with (1250, 2500] mentions.}
\label{fig:speed_comp}
\end{figure}

This section compares the computational performance of \texttt{longdoc} with the proposed \modelname-S architecture. The classification formulation and the lack of an update step in \modelname-S makes it a more efficient alternative to \modelname-H and CR models. \cref{fig:speed_comp} displays the speed-up obtained in the identification module when assessed across documents with varying numbers of mentions. \modelname-S consistently clocks a 20x efficiency across all ranges.

\subsection{Performance across number of entities}
\label{app:var_ent}
\begin{figure}[t]
\centering
\includegraphics[width=0.99\linewidth]{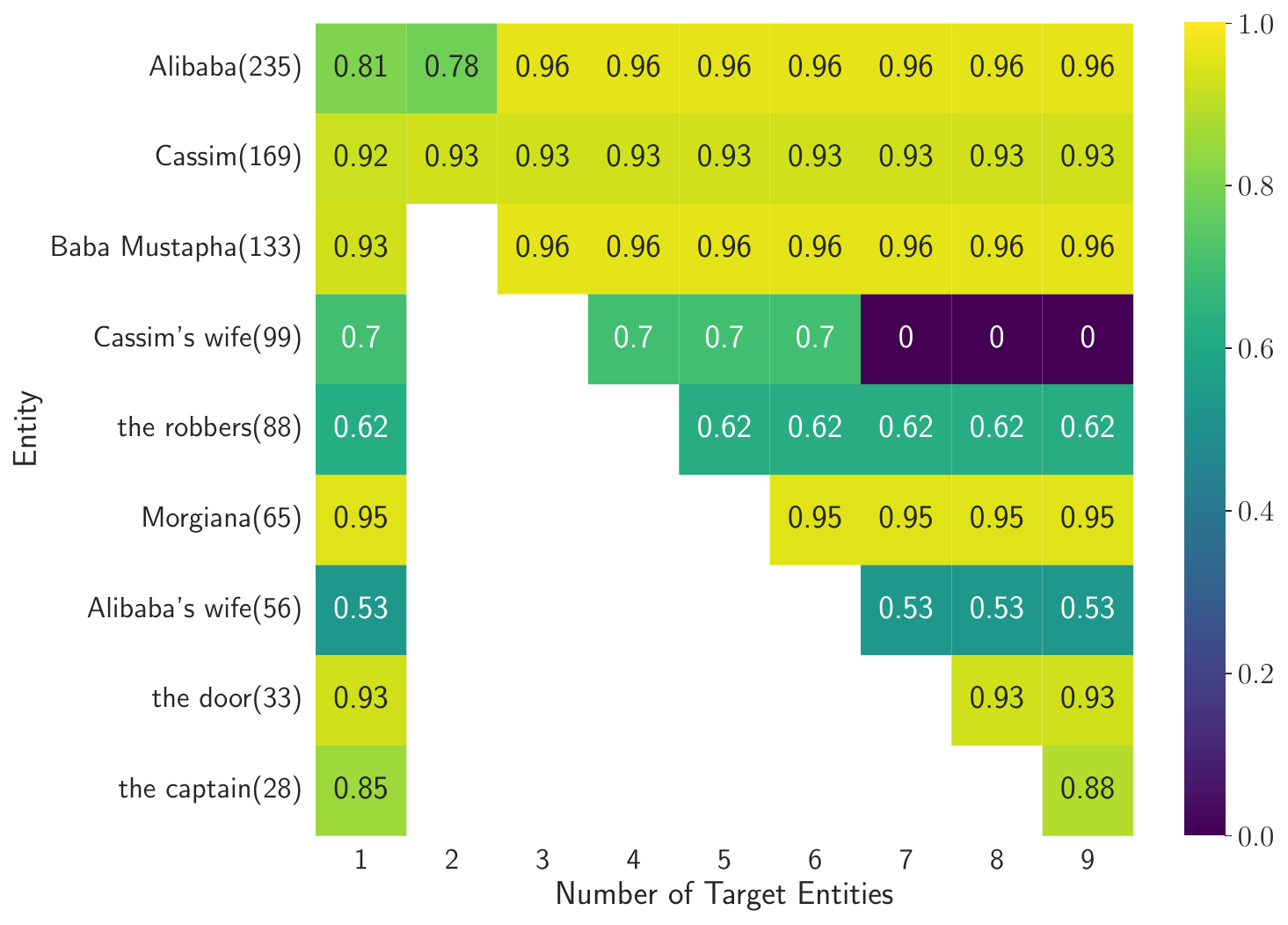}
\caption{Performance of \modelname-H across number of target entities for the document Ali Baba and the Forty Thieves.} 
\label{fig:alibaba}
\end{figure}

\begin{figure}[t]
\centering
\includegraphics[width=0.99\linewidth]{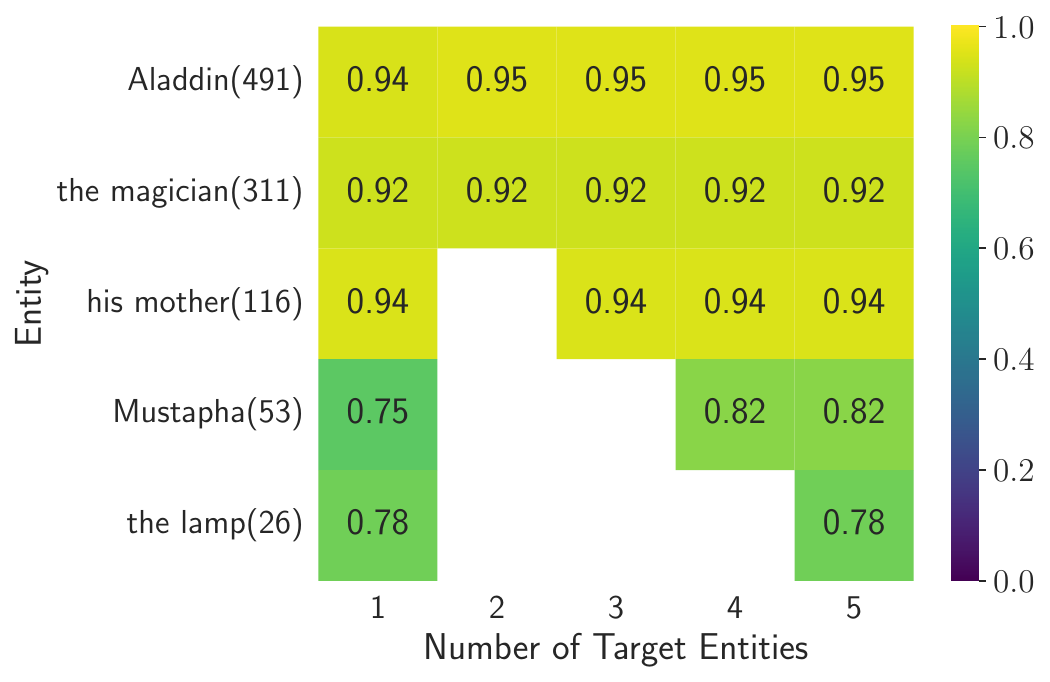}
\caption{Performance of \modelname-H across number of target entities for Aladdin.} 
\label{fig:aladdin}
\end{figure}

\begin{figure}[t!]
\centering
\includegraphics[width=0.99\linewidth]{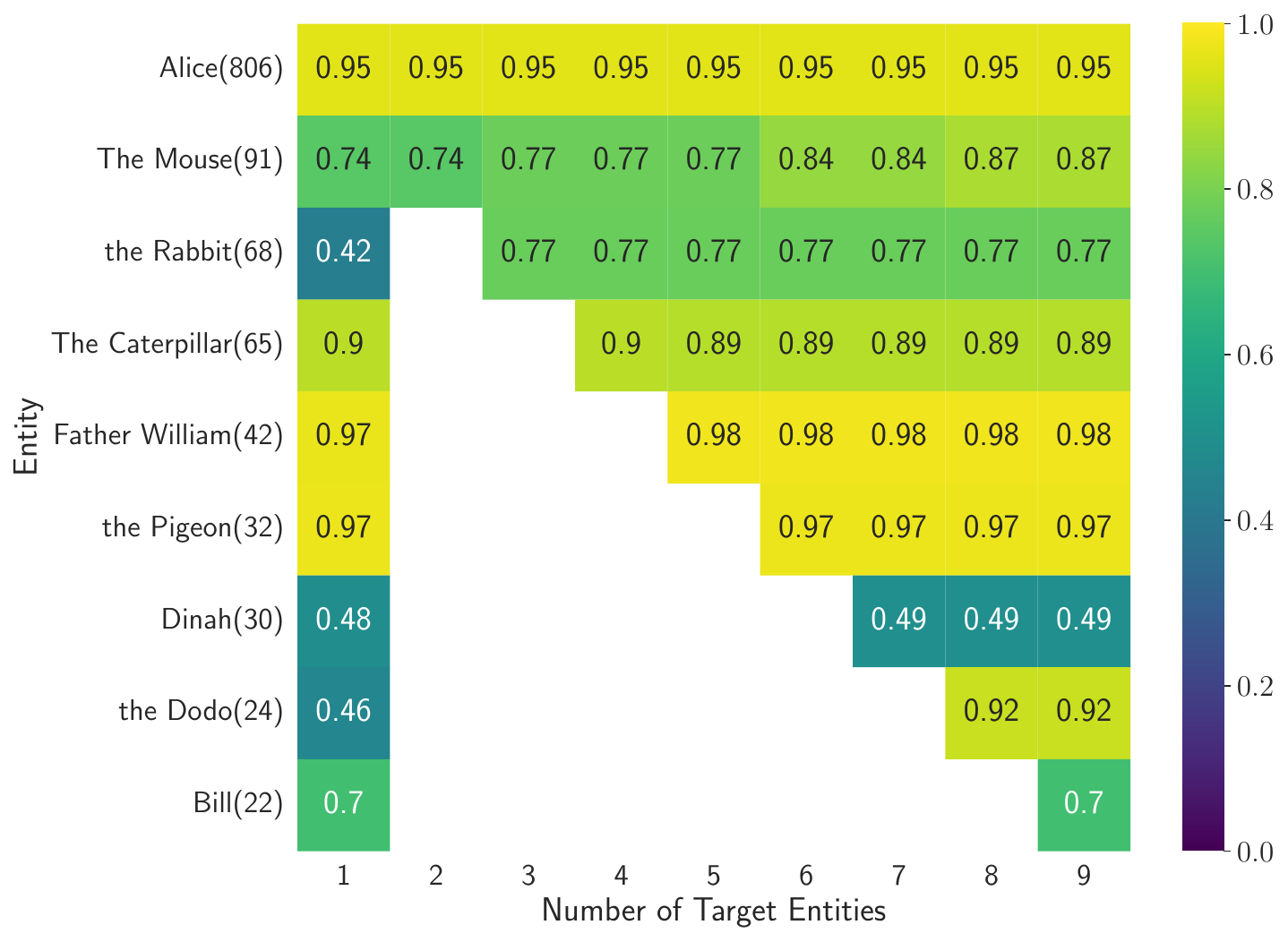}
\caption{Performance of \modelname-H across number of target entities for Alice in Wonderland.} 
\label{fig:alice}
\end{figure}

For consistency, the experiments of the main paper are evaluated across all the selected major entities (chosen using the thresholds defined in \cref{sec:experiments}). A natural extension is to assess the model's performance with varying numbers of entities of choice. For instance, if one is interested in only two key characters, can these models maintain consistency when provided with their designative phrases?

In this section, we address this concern and evaluate the MEI models with varying numbers of input entities. We present the per-entity F1-score of all entities across the AFT dataset. The results for \modelname-H are showcased in \cref{fig:alibaba}, \cref{fig:aladdin} and \cref{fig:alice}. The first column of the heatmap shows the per-entity F1-score when it is the sole target entity in the document. For \,  e.g., the value in the first column in \cref{fig:alibaba} corresponding to the entity \emph{Baba Mustapha} (0.93) indicates the performance of the model when \emph{Baba Mustapha} is the only target entity.

As we move across the columns of a particular row (ignoring the first column), the column number indicates the number of target entities used at inference. For instance, if the column number is $k$, the target entities are the top-$k$ frequent entities. Again, the 4\textsuperscript{th} column in the row corresponding to \emph{Baba Mustapha} indicates its individual F1-score in the experiment where the four input entities are \emph{Alibaba}, \emph{Cassim}, \emph{Baba Mustapha} and \emph{Cassim's wife}. 

There are a few individual cases where the performance significantly varies with modifying the number of input entities. For example, \emph{Cassim's wife} is confused with \emph{Alibaba's wife} after the latter's introduction. However, overall, the per-entity F1-score remains consistent across varying numbers of input entities across all three documents. These results demonstrate the effectiveness of \modelname-H for applications requiring variable numbers of target entities. This consistency is mainly due to the variable entity training, where a randomly chosen subset of major entities is selected in each iteration. Excluding this procedure leads to significant fluctuation in performance while modifying the number of target entities.

\subsection{Prompts}
\label{sec:appendix_prompts}
We provide exact prompts for all the few-shot prompting experiments. Please note that not all the major entities listed in the few shot examples are necessary to be present in the text. 

\begin{table*}[t!]
\centering
\small
\ttfamily
\begin{tabular}{p{15cm}}
\toprule
\textbf{Instruction} \\
\midrule
You will receive a Text along with a list of Key Entities and their corresponding Cluster IDs as input. Your task is to perform Coreference Resolution on the provided text to categorize "each word belonging to a cluster" with its respective cluster id. Also briefly describe the key entities in 1-2 sentences before starting the coreference task.\\

Follow the format below to label a word with its cluster ID: \\
word\#cluster\_id \\

Please keep in mind:\\
- Ensure the output adheres to the specified format for easy parsing. \\
- Classify the words in the given text without altering any of the other content. \\ \midrule
\textbf{Example Input:} \\ \midrule
Key Entities: \\
1. \color{red}Katharine Hilbery - \#1 \\
2. \color{green}Mr. Denham - \#2 \\
3. \color{blue}Mrs. Hilbery - \#3 \\
4. \color{brown}Mr. Hilbery - \#4 \\
5. \color{teal}Mr. Fortescue - \#5 \\
\\
Text: \\
CHAPTER I It was a Sunday evening in October , and in common with many other young ladies of her class , Katharine Hilbery was pouring out tea . Perhaps a fifth part of her mind was thus occupied , and the remaining parts leapt over the little barrier of day which interposed between Monday morning and this rather subdued moment , and played with the things one does voluntarily and normally in the daylight . But although she was silent , she was evidently mistress of a situation which was familiar enough to her , and inclined to let it take its way for the six hundredth time , perhaps , without bringing into play any of her unoccupied faculties . A single glance was enough to show that Mrs. Hilbery was so rich in the gifts which make tea-parties of elderly distinguished people successful , that she scarcely needed any help from her daughter , provided that the tiresome business of teacups and bread and butter was discharged for her . Considering that the little party had been seated round the tea-table for less than twenty minutes , the animation observable on their faces , and the amount of sound they were producing collectively , were very creditable to the hostess . It suddenly came into Katharine 's mind that if someone opened the door at this moment he would think that they were enjoying themselves ; he would think , " What an extremely nice house to come into ! " \\ \midrule
\textbf{Example Output:} \\ \midrule
Description of Key Entities present in the text: \\
\#1 - Katharine Hilbery: A young and apparently rich lady and the daughter of Mrs. Hilbery. She and Mrs. Hilbery were organising a party for some distinguished elders. \\
\#3 - Mrs. Hilbery: She is the mother of Katharine Hilbery and is a well-to-do member of the society and a very efficient and able hostess\\
\\
Coreference: \\
CHAPTER I It was a Sunday evening in October , and in common with many other young ladies of {\color{red}her\#1} class , {\color{red}Katharine\#1 Hilbery\#1} was pouring out tea . Perhaps a fifth part of {\color{red}her\#1} mind was thus occupied , and the remaining parts leapt over the little barrier of day which interposed between Monday morning and this rather subdued moment , and played with the things one does voluntarily and normally in the daylight . But although {\color{red}she\#1} was silent , {\color{red}she\#1} was evidently mistress of a situation which was familiar enough to {\color{red}her\#1} , and inclined to let it take its way for the six hundredth time , perhaps , without bringing into play any of {\color{red}her\#1} unoccupied faculties . A single glance was enough to show that {\color{blue}Mrs.\#3 Hilbery\#3} was so rich in the gifts which make tea-parties of elderly distinguished people successful , that {\color{blue}she\#3} scarcely needed any help from {\color{blue}her\#3} {\color{red}daughter\#1} , provided that the tiresome business of teacups and bread and butter was discharged for {\color{red}her\#1} . Considering that the little party had been seated round the tea-table for less than twenty minutes , the animation observable on their faces , and the amount of sound they were producing collectively , were very creditable to the {\color{blue}hostess\#3} . It suddenly came into {\color{red}Katharine\#1 's\#1} mind that if some one opened the door at this moment he would think that they were enjoying themselves ; he would think , " What an extremely nice house to come into ! " 
 \\ \bottomrule
\end{tabular}%
\caption{Prompt for WL Coreference}
\label{tab:app_wl}
\end{table*}

\begin{table*}[t]
\small
\ttfamily
\begin{tabular}{p{15cm}}
\toprule
\textbf{Instruction} \\ \midrule
Any word marked with \# is supposed to be the head of a noun phrase. Expand this head to contain determiner and adjective phrases. Do not remove or add new words while expanding. Stick to the format. \\ \midrule
\textbf{Example Input:} \\ \midrule
\textbf{Montraville\#} was a \textbf{Lieutenant\#} in the \textbf{army\#} : \textbf{Belcour\#} was his brother \textbf{officer\#} : they had been to take leave of their \textbf{friends\#} previous to their departure for \textbf{America\#} , and were now returning to \textbf{Portsmouth\#} , where the \textbf{troops\#} waited orders for embarkation  \\ \midrule
\textbf{Example Output:} \\ \midrule
Montraville \textbf{(Montraville)} was a Lieutenant \textbf{(a Lieutenant in the army)} in the army \textbf{(the army)} : Belcour \textbf{(Belcour)} was his brother officer \textbf{(his brother officer)} : they had been to take leave of their friends \textbf{(their friends)} previous to their departure for America \textbf{(America)} , and were now returning to Portsmouth \textbf{(Portsmouth)} , where the troops \textbf{(the troops)} waited orders for embarkation \\ \midrule
\textbf{Example Input:} \\ \midrule
Arriving at the verge of the \textbf{town\#} , he dismounted , and sending the \textbf{servant\#} forward with the horses , proceeded toward the \textbf{place\#} , where , in the midst of an extensive pleasure \textbf{ground\#} , stood the \textbf{mansion\#} which contained the lovely Charlotte \textbf{Temple\#} . \\ \midrule
\textbf{Example Output:} \\ \midrule
Arriving at the verge of the town \textbf{(the town)} , he dismounted , and sending the servant \textbf{(the servant)} forward with the horses , proceeded toward the place \textbf{(the place)} , where , in the midst of an extensive pleasure ground \textbf{(an extensive pleasure ground)} , stood the mansion \textbf{(the mansion which contained the lovely Charlotte Temple)} which contained the lovely Charlotte Temple \textbf{(the lovely Charlotte Temple)} . \\ \midrule
\textbf{Example Input:} \\ \midrule
"You are a benevolent \textbf{fellow\#} ," said a young \textbf{officer\#} to him one day and I have a great mind to give you a fine subject to exercise the goodness of your heart upon. \\ \midrule
\textbf{Example Output:} \\ \midrule
"You are a benevolent fellow \textbf{(a benevolent fellow)} ," said a young officer \textbf{(a young officer)} to him one day and I have a great mind to give you a fine subject to exercise the goodness of your heart upon. \\ \bottomrule
\end{tabular}%
\caption{Prompt for H2S Retrieval}
\label{tab:app_h2s}
\end{table*}

\begin{table*}[t]
\small
\ttfamily
\begin{tabular}{p{15cm}}
\toprule
\textbf{Instruction} \\ \midrule
Annotate all the entity mentions in the following text with coreference clusters. Use Markdown tags to indicate clusters in the output, with the following format [mention] (\#cluster\_name).
Do not modify any text outside (), only add text inside parenthesis. The cluster names of the key entities are already provided, mark the mentions of the entity with the corresponding cluster name. Mark the mentions of the other entities with (\#others). Also briefly describe the key entities in 1-2 sentences before starting the coreference task. \\ \midrule
\textbf{Example Input:} \\ \midrule
 Key Entities:\\
1. \color{red}Katharine Hilbery (\#katharine\_hilbery) \\
2. \color{green}Mr. Denham (\#mr.\_denham) \\
3. \color{blue}Mrs. Hilbery (\#mrs.\_hilbery) \\
4. \color{brown}Mr. Hilbery (\#mr.\_hilbery) \\
5. \color{teal}Mr. Fortescue (\#mr.\_fortescue) \\
\\
Text: \\
CHAPTER I It was a Sunday evening in October, and in common with [many other young ladies of [her] (\#) class] (\#) , [Katharine Hilbery] (\#) was pouring out tea . Perhaps a fifth part of [her] (\#) mind was thus occupied , and the remaining parts leapt over the little barrier of day which interposed between Monday morning and this rather subdued moment , and played with the things one does voluntarily and normally in the daylight . But although [she] (\#) was silent , [she] (\#) was evidently [mistress] (\#) of a situation which was familiar enough to [her] (\#) , and inclined to let it take its way for the six hundredth time , perhaps , without bringing into play any of [her] (\#) unoccupied faculties . A single glance was enough to show that [Mrs. Hilbery] (\#) was so rich in the gifts which make tea-parties of [elderly distinguished people] (\#) successful , that [she] (\#) scarcely needed any help from [[her] (\#) daughter] (\#) , provided that the tiresome business of teacups and bread and butter was discharged for [her] (\#) . Considering that [the little party] (\#) had been seated round the tea-table for less than twenty minutes , the animation observable on [their] (\#) faces , and the amount of sound [they] (\#) were producing collectively , were very creditable to [the hostess] (\#) . It suddenly came into [Katharine 's] (\#) mind that if [some one] (\#) opened the door at this moment [he] (\#) would think that [they] (\#) were enjoying [themselves] (\#) ; [he] (\#) would think , " What [an extremely nice house] (\#) to come into ! " \\ \midrule
\textbf{Example Output:} \\ \midrule
Description of Key Entities present in the text: \\
{\color{red}\#katharine\_hilbery - Katharine Hilbery}: A young and apparently rich lady and the daughter of Mrs. Hilbery. She and Mrs. Hilbery were organising a party for some distinguished elders. \\
{\color{blue}\#mrs.\_hilbery - Mrs. Hilbery}: She is the mother of Katharine Hilbery and is a well-to-do member of the society and a very efficient and able hostess\\
\\
Coreference: \\
CHAPTER I It was a Sunday evening in October , and in common with {\color{cyan}[many other young ladies of {\color{red}[her] (\#katharine\_hilbery)} class] (\#others)} , {\color{red}[Katharine Hilbery] (\#katharine\_hilbery)} was pouring out tea . Perhaps a fifth part of {\color{red}[her] (\#katharine\_hilbery)} mind was thus occupied , and the remaining parts leapt over the little barrier of day which interposed between Monday morning and this rather subdued moment , and played with the things one does voluntarily and normally in the daylight . But although {\color{red}[she] (\#katharine\_hilbery)} was silent , {\color{red}[she] (\#katharine\_hilbery)} was evidently {\color{cyan}[mistress] (\#others)} of a situation which was familiar enough to {\color{red}[her] (\#katharine\_hilbery)} , and inclined to let it take its way for the six hundredth time , perhaps , without bringing into play any of {\color{red}[her] (\#katharine\_hilbery)} unoccupied faculties . A single glance was enough to show that {\color{blue}[Mrs. Hilbery] (\#mrs.\_hilbery)} was so rich in the gifts which make tea-parties of {\color{cyan}[elderly distinguished people] (\#others)} successful , that {\color{blue}[she] (\#mrs.\_hilbery)} scarcely needed any help from {\color{red}[}{\color{blue}[her] (\#mrs.\_hilbery) }{\color{red} daughter] (\#katharine\_hilbery)} , provided that the tiresome business of teacups and bread and butter was discharged for {\color{red}[her] (\#katharine\_hilbery)} . Considering that {\color{cyan}[the little party] (\#others)} had been seated round the tea-table for less than twenty minutes , the animation observable on {\color{cyan}[their] (\#others)} faces , and the amount of sound {\color{cyan}[they] (\#others)} were producing collectively , were very creditable to {\color{blue}[the hostess] (\#mrs.\_hilbery)} . It suddenly came into {\color{red}[Katharine 's] (\#katharine\_hilbery)} mind that if {\color{cyan}[some one] (\#others)} opened the door at this moment {\color{cyan}[he] (\#others)} would think that {\color{cyan}[they] (\#others)} were enjoying {\color{cyan}[themselves] (\#others)} ; {\color{cyan}[he] (\#others)} would think , " What {\color{cyan}[an extremely nice house] (\#others)} to come into ! " \\ \bottomrule
\end{tabular}%
\caption{Prompt for evaluating linking performance}
\label{tab:app_link}
\end{table*}

\begin{table*}[t!]
\centering
\small
\ttfamily
\begin{tabular}{p{15cm}}
\toprule
\textbf{Instruction} \\
\midrule
Annotate all the entity mentions that refer to the key entities provided. The mention needs to include determiners and adjectives, if present. Use Markdown tags to indicate clusters in the output, with the following format [mention] (\#cluster\_name). The cluster names of the key entitites are already provided. Mark the mentions of the entity with the corresponding cluster name. Also briefly describe the key entities in 1-2 sentences before starting the coreference task.\\ \midrule
\textbf{Example Input:} \\ \midrule
Key Entities: \\
1. \color{red} Katharine Hilbery (\#katharine\_hilbery) \\
2. \color{green}Mr. Denham (\#mr.\_denham) \\
3. \color{blue}Mrs. Hilbery (\#mrs.\_hilbery) \\
4. \color{brown}Mr. Hilbery (\#mr.\_hilbery) \\
5. \color{teal}Mr. Fortescue (\#mr.\_fortescue) \\
\\
Text: \\
CHAPTER I It was a Sunday evening in October , and in common with many other young ladies of her class , Katharine Hilbery was pouring out tea . Perhaps a fifth part of her mind was thus occupied , and the remaining parts leapt over the little barrier of day which interposed between Monday morning and this rather subdued moment , and played with the things one does voluntarily and normally in the daylight . But although she was silent , she was evidently mistress of a situation which was familiar enough to her , and inclined to let it take its way for the six hundredth time , perhaps , without bringing into play any of her unoccupied faculties . A single glance was enough to show that Mrs. Hilbery was so rich in the gifts which make tea-parties of elderly distinguished people successful , that she scarcely needed any help from her daughter , provided that the tiresome business of teacups and bread and butter was discharged for her . Considering that the little party had been seated round the tea-table for less than twenty minutes , the animation observable on their faces , and the amount of sound they were producing collectively , were very creditable to the hostess . It suddenly came into Katharine 's mind that if someone opened the door at this moment he would think that they were enjoying themselves ; he would think , " What an extremely nice house to come into ! " \\ \midrule
\textbf{Example Output:} \\ \midrule
Description of Key Entities present in the text: \\
{\color{red}\#katharine\_hilbery - Katharine Hilbery}: A young and apparently rich lady and the daughter of Mrs. Hilbery. She and Mrs. Hilbery were organising a party for some distinguished elders. \\
{\color{blue}\#mrs.\_hilbery - Mrs. Hilbery}: She is the mother of Katharine Hilbery and is a well-to-do member of the society and a very efficient and able hostess\\
\\
Coreference: \\
CHAPTER I It was a Sunday evening in October , and in common with many other young ladies of {\color{red}[her] (\#katharine\_hilbery)} class , {\color{red}[Katharine Hilbery] (\#katharine\_hilbery)} was pouring out tea . Perhaps a fifth part of {\color{red}[her] (\#katharine\_hilbery)} mind was thus occupied , and the remaining parts leapt over the little barrier of day which interposed between Monday morning and this rather subdued moment , and played with the things one does voluntarily and normally in the daylight . But although {\color{red}[she] (\#katharine\_hilbery)} was silent , {\color{red}[she] (\#katharine\_hilbery)} was evidently mistress of a situation which was familiar enough to {\color{red}[her] (\#katharine\_hilbery)} , and inclined to let it take its way for the six hundredth time , perhaps , without bringing into play any of {\color{red}[her] (\#katharine\_hilbery)} unoccupied faculties . A single glance was enough to show that {\color{blue}[Mrs. Hilbery] (\#mrs.\_hilbery)} was so rich in the gifts which make tea-parties of elderly distinguished people successful , that {\color{blue}[she] (\#mrs.\_hilbery)} scarcely needed any help from {\color{red}[{\color{blue}[her] (\#mrs.\_hilbery)} daughter] (\#katharine\_hilbery)} , provided that the tiresome business of teacups and bread and butter was discharged for {\color{red}[her] (\#katharine\_hilbery)} . Considering that the little party had been seated round the tea-table for less than twenty minutes , the animation observable on their faces , and the amount of sound they were producing collectively , were very creditable to {\color{blue}[the hostess] (\#mrs.\_hilbery)} . It suddenly came into {\color{red}[Katharine 's] (\#katharine\_hilbery)} mind that if some one opened the door at this moment he would think that they were enjoying themselves ; he would think , " What an extremely nice house to come into ! " \\ \bottomrule
\end{tabular}%
\caption{Prompt for Direct version of E2E MEI}
\label{tab:app_dir}
\end{table*}

\subsection{Budget and Hardware details}
The supervised models were trained on a 24GB NVIDIA RTX 4090Ti GPU. For experiments with the open source language models, we used two 48GB NVIDIA RTX A6000 GPU's. For GPT-4 and GPT-3.5 experiments, we spent approximately 175\$ in total, covering both initial explorations and the computation of final results.

\end{document}